\documentclass[runningheads]{llncs}
\usepackage{hyperref}

\usepackage{amsmath}
\usepackage{amssymb}
\usepackage{booktabs}
\usepackage{cite}
\usepackage{picinpar}
\usepackage{url}
\usepackage{flushend}
\usepackage[latin1]{inputenc}

\usepackage{colortbl}
\usepackage{soul}
\usepackage{multirow}
\usepackage{pifont}
\usepackage{alltt}
\usepackage{enumerate}
\usepackage{siunitx}
\usepackage{breakurl}
\usepackage{epstopdf}
\usepackage{pbox}
\usepackage{amsfonts}
\usepackage{diagbox}
\usepackage{graphicx}

\usepackage{tikz}
\usepackage{comment}
\usepackage{amsmath,amssymb} 
\usepackage{color}

\usepackage[accsupp]{axessibility}  

\begin{document}
\pagestyle{headings}
\mainmatter
\def\ECCVSubNumber{5544}  

\title{An Efficient Person Clustering Algorithm for Open Checkout-free Groceries} 

\titlerunning{An Efficient Person Clustering Algorithm for Open Checkout-free Groceries}
%
\author{Junde Wu\inst{1,}*$^{,\dag}$  \and 
Yu Zhang\inst{2,}* \and 
Rao Fu\inst{2} \and 
Yuanpei Liu\inst{3} 
\and Jing Gao \inst{1} \\~\\
* equal technical contribution \qquad $^{\dag}$ project lead
}
\authorrunning{J. et al.}
%
\institute{Purdue University, West Lafayette \and Harbin Institute of Technology
\and The University of Hong Kong
}
\maketitle

\begin{abstract}
Open checkout-free grocery is the grocery store where the customers never have to wait in line to check out. Developing a system like this is not trivial since it faces challenges of recognizing the dynamic and massive flow of people. In particular, a clustering method that can efficiently assign each snapshot to the corresponding customer is essential for the system. In order to address the unique challenges in the open checkout-free grocery, we propose an efficient and effective person clustering method. 
Specifically, we first propose a Crowded Sub-Graph (CSG) to localize the relationship among massive and continuous data streams. CSG is constructed by the proposed Pick-Link-Weight (PLW) strategy, which  \textbf{picks} the nodes based on time-space information, \textbf{links} the nodes via trajectory information, and \textbf{weighs} the links by the proposed von Mises-Fisher (vMF) similarity metric. Then, to ensure that the method adapts to the dynamic and unseen person flow, we propose Graph Convolutional Network (GCN) with a simple Nearest Neighbor (NN) strategy to accurately cluster the instances of CSG. GCN is adopted to project the features into low-dimensional separable space, and NN is able to quickly produce a result in this space upon dynamic person flow. The experimental results show that the proposed method outperforms other alternative algorithms in this scenario. In practice, the whole system has been implemented and deployed in several real-world open checkout-free groceries. 
{We publicly release a collected large-scale real-world check-out free grocery dataset at:  \href{https://github.com/WuJunde/checkoutfree}{https://github.com/WuJunde/checkoutfree}.}
\end{abstract}

\section{Introduction}
Traditional checkout-free groceries are in small closed venues with limited commodities. Customers are required to register upon entry, which may cause privacy and security issues. 
Recently, a concept of open checkout-free grocery is proposed to address the existing issues. Open checkout-free grocery allows free entry of consumers without registering customer information. Customers can walk in such grocery stores just like what they do in the traditional supermarkets while enjoying the benefit of automatic checkout. To achieve this goal, it is essential to automatically identify the customers in the grocery, which is a very challenging task in an open environment. The number of needed identifications is not accessible beforehand, and the identification process needs to work continuously and steadily for every customer. Therefore, many tools in computer vision to associate/track a person, e.g., human tracking, person re-identification, or face verification, cannot be directly applied in this scenario. Instead, a person clustering component is essential to cluster the customers based on features extracted from their video snapshots. Implementing an effective clustering algorithm in this scenario is non-trivial. Among many challenges, below we discuss the two major challenges of this problem. 

\begin{figure}[h!]
\centering
  \includegraphics[width=0.85\linewidth]{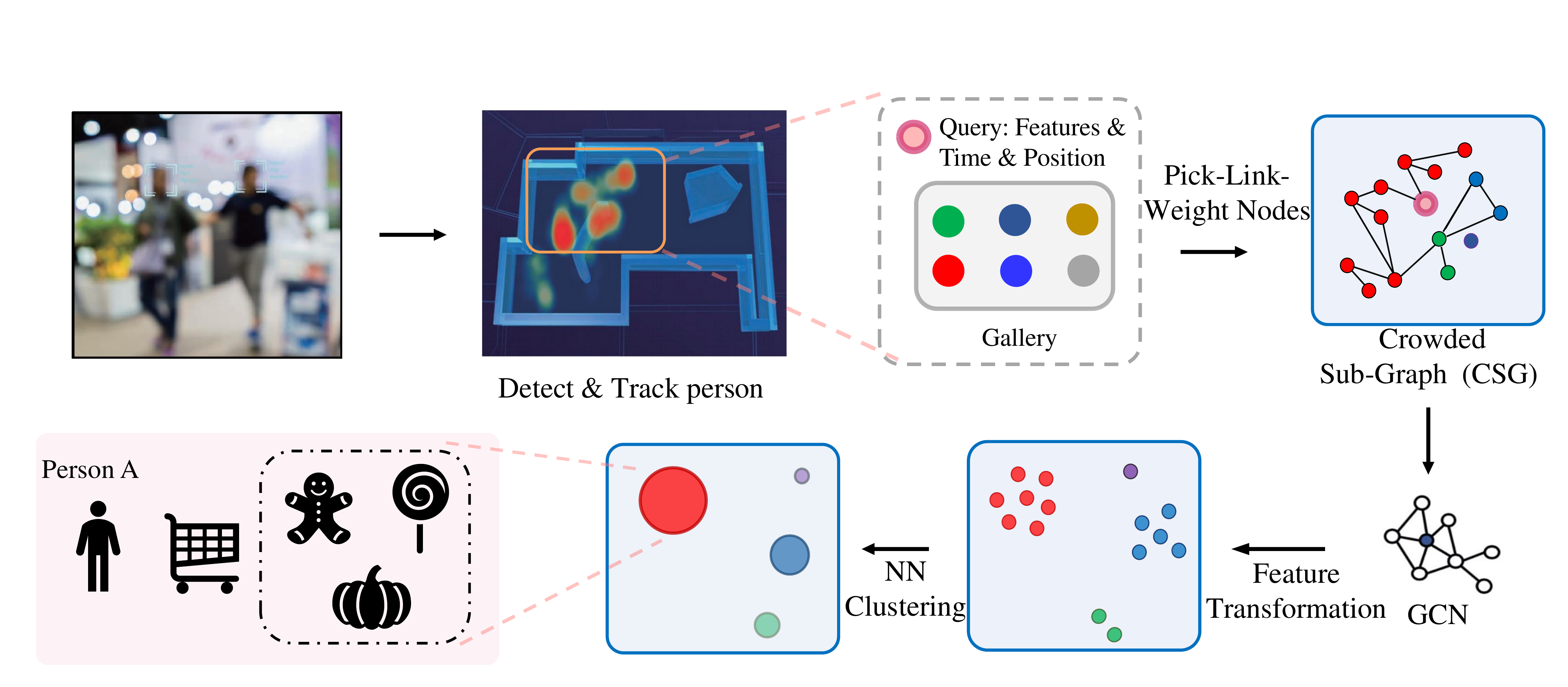}
  \caption{ Overview of the person clustering algorithm in open checkout-free groceries. 
  }
  \label{fig:fig1}
\end{figure}

The first challenge comes from the difficulty of handling data streams\cite{aggarwal2007data}, which are the input of the person clustering algorithm. The proposed clustering algorithm has to work over massive, unbounded sequences of data objects that are continuously generated at rapid rates. In addition, the algorithm must react immediately when a \textit{query} comes, under an extremely harsh condition where the observed data stream is generally too large to store and too expensive to access.

The second challenge is that person clustering in this scenario is an open-world problem \cite{bendale2015towards}. The data that the system needs to process is dynamic and unknown person flow. Therefore, the system is required to continuously distinguish and memorize the unseen new customer once an individual comes and could recognize the person in any other locations and views since then. 


For the first challenge, the commonly applied method is to cluster locally on the summarized data (statistic summaries of data streams) \cite{aggarwal2007data}. In this paper, we propose a Crowded Sub-Graph (CSG) to efficiently collect the local clusters and provide the divergence of the summarized data. CSG constructs a local relationship network for each incoming \textit{query} through the Pick-Link-Weight (PLW) strategy. Specially, we leverage time-space information to \textbf{pick} the nodes, and utilize trajectory information to \textbf{link} the nodes, and propose a novelty von Mises-Fisher (vMF) similarity metric to \textbf{weigh} the links. With the CSG, we can compare locally to accelerate the system without degrading performance.

The open-world nature of the scenario leads to dynamic changes in the distribution of person features. 
Thus, some complex clustering techniques may consume too much time, while some simple techniques may lead to performance degradation. In this paper, we propose Graph Convolutional Network (GCN) with a simple Nearest Neighbor (NN) strategy to accurately cluster the instances of CSG. GCN is adopted to project the features into low-dimensional separable space, and NN is able to quickly produce a result in this space upon dynamic person flow. To our knowledge, our work is the first towards a comprehensive strategy to identify a person in this data-stream \& open-world environment.

In brief, the contributions of the paper are as follows. 
\begin{itemize}
\item We are the first to define the People Clustering task in an open checkout-free grocery scenario, and we propose an effective framework to address this important problem. As the first research report of this scene, we believe it will be an essential starting point for future research and practical applications. 
\item We propose Crowded Sub-Graph (CSG), a local relationship graph constructed by PLW strategy. PLW strategy can model the distance of nodes through the lens of the probability distribution, so as to construct a sub-graph that fairly represents the relevance of local nodes.
\item Given CSG as input data, we apply Graph Convolutional Network (GCN) on it following a simple NN (Nearest Neighbor) strategy to cluster the nodes, which is able to quickly adapt to the dynamic person distribution in groceries. Experimental results show the proposed algorithm considerably outperforms best alternative methods.
\end{itemize}
\section{Related Works}
\subsection{Data-stream clustering}
For the last decade, we have seen an increasing interest in managing the data streams \cite{aggarwal2007data, gama2010knowledge}. Clustering on data streams requires a process to continuously cluster data objects within memory and time restrictions \cite{gama2010knowledge}.

In the data abstraction step, the data structures to summarize the data are also diversely proposed for the different tasks\cite{silva2013data}, like feature vectors \cite{zhang1996birch}, prototype arrays \cite{domingos2001general}, coreset trees \cite{ackermann2012streamkm++}, and data grids \cite{kranen2011clustree}. However, most of them are bound with particular clustering methods, which narrows their applications.

In the clustering step, many k-means variants have been presented to deal with summarized features to cluster in data streams in real-time. Bradley et al. \cite{bradley1998scaling} proposed Scalable k-means, which uses the CF vectors of the processed and new data objects as input to find $k$ clusters. 
The ClusTree algorithm \cite{kranen2011clustree} proposes to use a weighted CF vector, which is kept into a hierarchical tree (R-tree family). These well-designed data-stream clustering methods, on the one hand, to date, are limited to Euclidean spaces, and on the other hand, hard to take a balance between efficiency and performance.
\subsection{GCN on clustering}
Recently, with a part of the various applications of deep neural networks\cite{wu2020leveraging, wu2022learning, wu2022opinions, wu2022seatrans, ji2021learning}, Graph Convolutional Network (GCN) has shown outstanding performance on data clustering. GCN can extract high-level node representations, thus simplifying the sensitive discrimination step\cite{Wu2019UniversalTA}.

To apply GCN on the data which naturally has the graph structure seems straightforward, such as graph-based recommendation systems \cite{tang2021dynamic}, point clouds classification \cite{PointviewGCN}, and molecular properties prediction \cite{ryu2019bayesian}, etc. However, the graph nature of some other data may not be so explicit. In this case, researchers have to construct the graph of the data. For example, 
\cite{zhang2018gaan} addressed the traffic prediction problem using STGNNs. \cite{velivckovic2017graph} applied the GCN to text classification based on the syntactic dependency tree of a sentence. \cite{wang2019linkage} proposed Instance Pivot Subgraph (IPS) to construct the sub-graph for person face features. However, despite the outstanding performance of GCN on data clustering, there is still a research gap between GCN and data-stream clustering.
\section{Preliminary}
\subsection{Data-stream clustering}
A data stream $S$ is a massive sequence of data objects $x_{1}, x_{2},..., x_{K}$, that is, $S = \{x_{i}\}^{K}_{i=1}$, which is potentially unbounded ($K \to \infty$). Each data object is described by a n-dimensional attribute vector $x_{i} = [x_{i}^{j}]^{n}_{j=1}$ belonging to an attribute space $\Omega$ that can be continuous, categorical, or mixed. 

It is impossible to store and get access to each data object in the data stream. Developing suitable data structures to store statistic summaries of data streams is indispensable in the data-stream clustering tasks. Cluster Feature vector (CF vector) is a commonly used data structure for summarizing large amounts of data. The CF vector has three components: $K$, the number of data objects, $LS$, the linear sum of the data objects, and $SS$, the sum of squared data objects. The structures $LS$ and $SS$ are n-dimensional vectors. These three components allow to compute cluster measures, such as cluster mean $\mu$ and radius $\sigma$ (Eq. (\ref{centroid})).
\begin{equation}\label{centroid}
    \mu = \frac{LS}{K}, \  \sigma = \sqrt{\left( \frac{SS}{K}-\left( \frac{LS}{K} \right) ^{2} \right)} ,
\end{equation}
\noindent where $(\cdot)^{2}$ and $\sqrt{\cdot}$ represent element-wise square and square root. Obviously, the three components of the CF vector have incrementality and additivity properties, which make the CF vector widely used in clustering (More details can be found in the supplementary material, or \cite{silva2013data}).

In the open checkout-free groceries, a complete \textit{person record} $p$ is represented as a set (Eq. (\ref{person_record1}))
\begin{equation}\label{person_record1}
p = \{ \{\tilde{t}_{i} \}_{i = 1}^{K},\{ z_{i} \}_{i = 1}^{K},\{ v_{i} \}_{i = 1}^{K}, CF[K]\},
\end{equation}
where $\tilde{t}_{i} = t_{i}^{s} + t_{i}^{e}$, $t_{i}^{s}$ and $t_{i}^{e}$ are the time stamps of the person appeared in the camera view and left the camera view, respectively. $z_{i}$ is a two dimensional point records the plane coordinates of the camera, $v_{i}$ is a two-dimensional normalized vector to denote the direction of the pedestrian's walking. It has $v_{i} = (\cos(\theta),\sin(\theta))$, where $\theta$ is the angle between the last straight pedestrian path in the camera and the horizontal line. The pedestrian path is got from the move path of the bounding box. $CF[K]$ represents incremented CF vectors of person features updated $K$ times. $K \geq 1$ is the number of pieces of data incremented in $p$. Each coming data was a \textit{query}, which is represented as:
\begin{equation}\label{person_record}
q = \{\tilde{t}, z,v, CF[1]\},
\end{equation}
where $CF[1]$ represents the initial tracked person features.
\section{Methodology}
The complete abstracted features in open checkout-free grocery combined several different sensors, and the vision system is one of the most important parts. The visual features are obtained through a flow of object tracking \cite{shen2021distilled}, person detection \cite{8634919}, image deblurring\cite{wu2020integrating}, image enhancement\cite{zhang2022better} and deep learning-based person feature abstraction \cite{almasawa2019survey}. When the camera captures a customer, it tracks the customer until the individual is out of view. We would sample several frames from this track and use a person detection algorithm to abstract a set of images of the person. These pictures will then be sent to the pre-trained neural networks to abstract a set of visual features. These visual features, combined with the appeared time-space information and person walking track, will be sent to the clustering algorithm to be identified. We call this piece of data sent to the clustering algorithm a \textit{query}.

The overflow of our algorithm contains two steps, which are shown in Fig. \ref{fig:fig1}. In the first step, when a \textit{query} comes, we construct a Crowded Sub-Graph for this \textit{query}. CSG contains N nodes, and one node is the \textit{query}, and the other N-1 nodes are \textit{person records}. The N-1 \textit{person records} are \textbf{picked} depending on the time-space constraint to assume a person is impossible to appear in a far place in a short time. Then we \textbf{link} the nodes depending on the person walking track and \textbf{weight} these links based on the vMF divergence of the person features. In the second step, we propose Graph Convolutional Network (GCN) with a simple Nearest Neighbor (NN) strategy to accurately cluster the instances of CSG. GCN is adopted to project the features into low-dimensional separable space, and NN is able to quickly produce a result in this space upon dynamic person flow. 
\subsection{Crowded Sub-Graph}
In this section, we introduce Crowded Sub-Graph (CSG) to construct a graph based on the raw data of open checkout-free groceries. Constructing a CSG contains three steps: to \textbf{pick} the nodes, \textbf{link} the nodes, and \textbf{weight} the links. In this way, CSG constructed a local sub-graph for the \textit{query}. The association of each pair of nodes in the sub-graph can be well represented.
\begin{figure}[h]
\centering
  \includegraphics[width=0.7\linewidth,height=3cm]{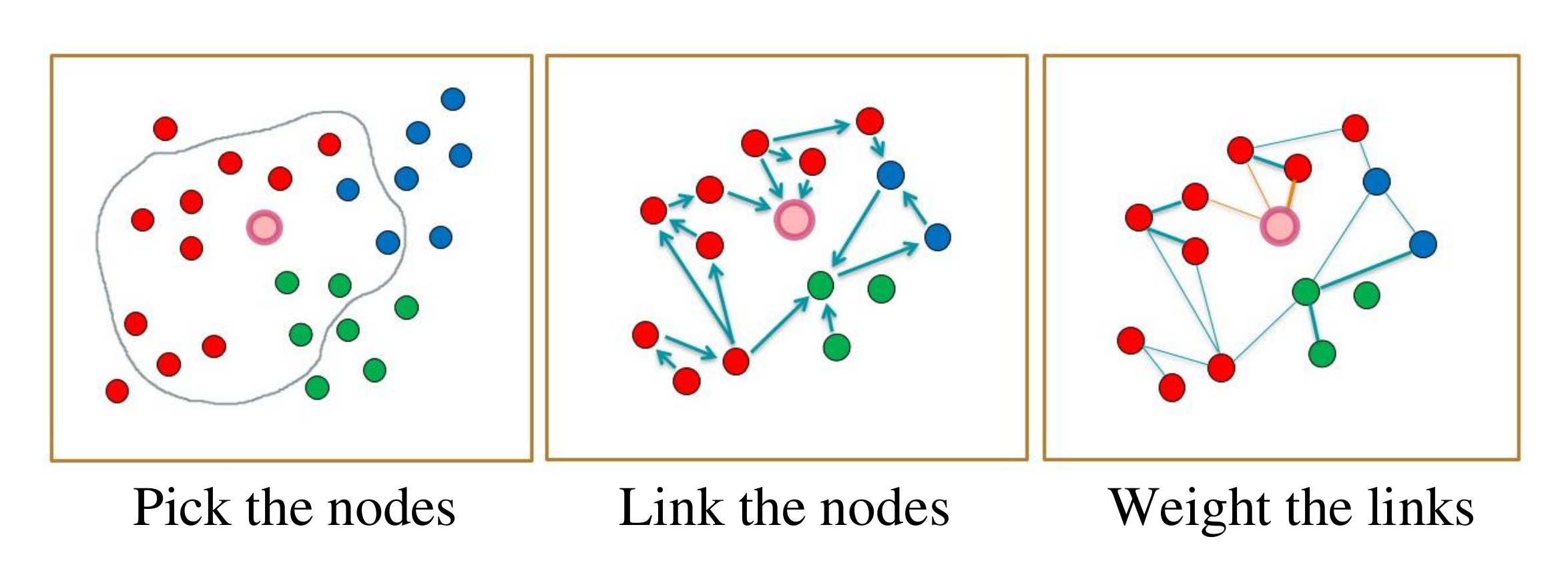}
  \caption{Construct Crowded Sub-Graph through the proposed Pick-Link-Weight (PLW) strategy}
  \label{fig:csg}
\end{figure}
\subsubsection{Pick the nodes}
Our algorithm is required to react immediately once a \textit{query} comes. However, clustering globally on the nearly unbounded data stream, i.e., clustering the \textit{query} based on all of the recorded person in the grocery, is impossible. Fortunately, a customer would not move far within the two captures of the cameras. This allows us to cluster locally based on time and space constraints. Considering a \textit{query} $q$ captured by a camera at location $z^{q}$ in time $t^{q}$, other \textit{person records} that have a smaller time-space distance with the \textit{query} $q$ would have a higher possibility to be contained to the sub-graph. For a \textit{person record} $p$ with time and location sets $\{\tilde{t}^{p}_{i} \}_{i = 1}^{K^{p}}$ and $\{ z^{p}_{i} \}_{i = 1}^{K^{p}}$, we define the time-space distance between $p$ and $q$ is:
\begin{equation}\label{E:dtspq}
    \chi (p;q) = \min \ \sqrt{\frac{E(z_{i}^{p},z^{q})^2}{s^2} + (\tilde{t}_{i}^{p} - \tilde{t}^{q})^2}, \ i \in [1,K^{p}],
\end{equation}
where $E$ denotes Euclidean distance, $s$ is the standard human walking speed, which is set as 3 miles per hour. Then we can collect the nodes of CSG. The number of the nodes we collect depends on the setting size of CSG.
\subsubsection{Link the nodes}
In the graph, two linked nodes usually mean that they are related to each other to some extent. In the open checkout-free grocery scenario, we assume two nodes are linked if they are on the same trajectory. Practically, we adopt an attentive dot-product mechanism \cite{vaswani2017attention} for the classification of the trajectories.  

\begin{figure}[h]
\centering
  \includegraphics[scale=0.32]{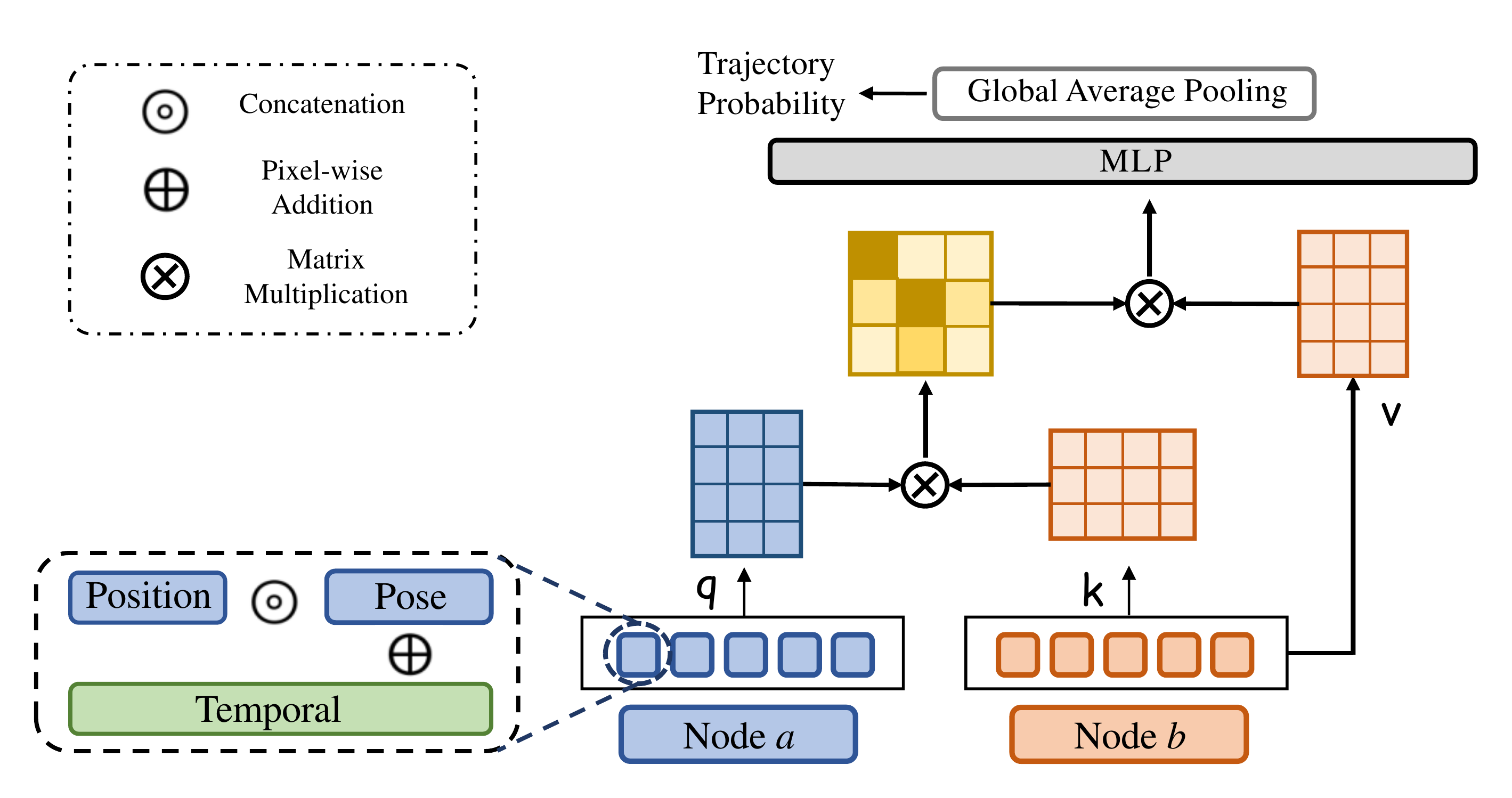}
  \caption{Trajectory predictor which predicts whether two nodes are on the same trajectory.}
  \label{fig_transformer}
\end{figure}

As shown in Fig. \ref{fig_transformer}, the recorded information of each pair of picked nodes are treated as inputs of the attention mechanism, including the position $z_i$ and pose $v_i$ of person relative to cameras and time stamps $t_i$. To predict whether Node $a$ is on the trajectory of node $b$, a QKV dot-product is adopted to activate node $b$ value matrix based on the calculated affinity map of two nodes. A  binary classification head is applied after the attention to get the probability of the prediction. The binary classification head consists of a Global Average Pooling (GAP) layer and the Multilayer Perceptron (MLP) layer. For each piece of recorded information in the node, we use the position vector, pose vector, and timestamp to constitute the input embedding. Specially, we concatenate the position vector and pose vector, then add a temporal embedding on it. The temporal embedding is learned from the time stamp, followed by \cite{carion2020end}. Formally, the sequential input embedding of trajectory predictor is represented as:
\begin{equation}\label{E:track1}
    [z_{i};v_{i}] + TE (t_{i}), i \in [1, K_{1} + K_{2}], 
\end{equation}
where $K_{1}$ and $K_{2}$ are the numbers of recorded information of two nodes. $TE(\cdot)$ represents the temporal embedding. We apply a trajectory predictor on each pair of nodes to get the final link relationship of the sub-graph. 
\subsubsection{Weight the link \& vMF similarity of CF vectors}
After we \textbf{link} the nodes in the graph, we then \textbf{weight} these links by measuring how strong the linked pair is associated. Specially, we propose a similarity distance for the CF vectors based on vMF distribution and use this distance to generate the weights.

Note that each node in CSG is represented as a CF vector. Since the centroid and radius of the node can be easily computed from CF vector, previous work often represented the distance of two nodes by the discrepancy of two normal distributions.
For example, a probabilistic model on the CF vector can be represented as a Gaussian model $\mathcal{N}(\mu,\sigma^{2})$, where mean $\mu$ and variance $\sigma$ can be easily through Equation (\ref{centroid}). Then, we can \textbf{weight} the link of node $a$ and node $b$ through the distance of $\mathcal{N}(\mu_{a},\sigma_{a}^{2})$ and $\mathcal{N}(\mu_{b},\sigma_{b}^{2})$.

However, such a practice implicitly assumes the distance of the features can be fairly represented by Euclidean distance, which is invalid for the high-dimensional person features in our hand. The person features abstracted by the neural networks have intrinsic angular distribution because of the softmax loss in the neural networks \cite{fan2019spherereid}. Thus the probability distributions on Euclidean space, like Gaussian distribution, are invalid on this $d$-sphere. In this paper, we use von Mises$-$Fisher (vMF) distribution instead to model these clusters \cite{banerjee2005clustering}. The von Mises$-$Fisher distribution is an isotropic distribution over the $d$-dimensional unit hypersphere, which can fairly represent the distribution high-dimensional person feature. It has mean direction $\mu$ and concentration $\kappa$, and the probability density function of it for the d-dimensional unit vector x is given by:
\begin{equation}
     f(\mu ,\kappa )=C_{d}(\kappa ) e^{\kappa \mu^{T} {x}},
\end{equation}
where $f(\cdot)$ represents the probability distribution, $\kappa \geq 0 $ , $\parallel \mu \parallel = 1$, and the normalization constant $ C_{p}(\kappa )$, is equal to
\begin{equation}
        C_{d}(\kappa )={\frac {\kappa ^{d/2-1}}{(2\pi )^{d/2}I_{d/2-1}(\kappa )}} ,
\end{equation}
where $ I_{v}$ denotes the modified Bessel function of the first kind at order v. In vMF distribution, $\mu$ denotes the mean direction, and $\kappa$ denotes the concentration. The greater the value of $\kappa$, the higher the concentration of the distribution around the mean direction $\mu$. When it is needed, we can follow \cite{sra2012short} to get the numerical solutions of $\kappa$ and $I_{v}$.

Consider two vMF probability distributions of Node $a$ and Node $b$ are $f(\theta_{a})$ and $f(\theta_{b})$,  and $f(\theta_{ab})$ denotes vMF distribution of their merged features. We \textbf{weight} the link $l_{ab}$ by:
\begin{equation}
    e^{-\frac{1}{2}(D_{JS}[f(\theta_{a}),f(\theta_{ab})] + D_{JS}[f(\theta_{b}),f(\theta_{ab})])},
\end{equation}
where $D_{JS}$ denotes Jensen$-$Shannon divergence \cite{fuglede2004jensen}:
\begin{equation}
\label{eqn12}
D_{JS}[f(\theta_{a}),f(\theta_{ab})] = \frac{1}{2}(D_{KL}(f(\theta_{a})) \| f(\theta_{ab}) + D_{KL}(f(\theta_{ab})) \| f(\theta_{a})),
\end{equation}
where $D_{KL}$ denotes Kullback$-$Leibler divergence.

To get the analytic solution of the KL divergence of two vMF distribution is challenging. For the benefit of the computation and graph sparsification, we \textbf{weight} the links of similar feature distributions and set the others as zero. We consider the parameters of a distribution are close to one another, so that:
\begin{equation}
f(\theta_{ab})=f(\theta _{a})+\sum _{j}\Delta \theta ^{j}\left.{\frac {\partial f}{\partial \theta ^{j}}}\right|_{\theta _{a}},
\end{equation}
where $\theta^{j}$ represents a small change of $ \theta $ in the $j$ direction. Then Kullback$-$Leibler divergence $ D_{\mathrm {KL} }[v(\theta _{a})\|v(\theta_{b})]$ has the second order Taylor Expansion in $\theta =\theta _{0}$ of the form
\begin{equation}\label{equation:dkl}
     D_{\mathrm {KL} }[f(\theta _{a})\|f(\theta_{ab})]={\frac {1}{2}}\sum _{jk}\Delta \theta ^{j}\Delta \theta ^{k} g_{jk}(\theta _{a})+\mathrm {O} (\Delta \theta ^{3}),
\end{equation}
in which
\begin{equation}\label{equation:fjk}
g_{jk}(\theta )=\int _{X}{\frac {\partial \log f(\theta )}{\partial \theta _{j}}}{\frac {\partial \log f(\theta )}{\partial \theta _{k}}}f(\theta )\,dx.  
\end{equation}
\par In our case, we have the parameter $\theta = (\kappa,\mu^{T})^{T}$. Substituting Eqn. (\ref{equation:fjk}) for the given parameters, we can get
\begin{equation}
    \begin{split}
    g_{\kappa,\kappa}(\kappa,\kappa) =& \tau_{\kappa} (\kappa,\mu), \ g_{\kappa,\mu}(\kappa,\mu) = \tau_{\kappa\mu} (\kappa)\mu^{T}, \\
    g_{\mu,\kappa}(\mu,\kappa) =& \tau_{\kappa\mu} (\kappa)\mu, \ g_{\mu,\mu}(\mu,\mu) = \tau_{\mu} (\kappa) \mu \mu^{T},
    \end{split}
\end{equation}
in which
\begin{equation}
    \begin{split}
    \tau_{\kappa}(\kappa,\mu) =&
[e^{\kappa\mu^{T}x}(\kappa \mu^{T} - 1)]^{2} + [\frac{d-2}{\kappa} - \frac{(I_{\frac{d}{2}-2}(\kappa) + I_{\frac{d}{2}}(\kappa))}{I_{\frac{d}{2}-1}(\kappa)}]  \\
    +&x[e^{\kappa\mu^{T}x}(\kappa \mu^{T} - 1)] + [\frac{d}{2} - \frac{2(I_{\frac{d}{2}-2}(\kappa) + I_{\frac{d}{2}}(\kappa))}{\kappa I_{\frac{d}{2}-1}(\kappa)} -1]^{2},
    \end{split}
\end{equation}
\begin{equation}
\begin{split}
            \tau_{\kappa\mu} (\kappa) =& \frac{(d-2)I_{\frac{d}{2}-1}(\kappa) -\kappa(I_{\frac{d}{2}-2}(\kappa) - I_{\frac{d}{2}}(\kappa))}{2(I_{\frac{d}{2}-1}(\kappa))^{2}},
\end{split}
\end{equation}
\begin{equation}
        \tau_{\mu} (\kappa) = \kappa^{2}.
\end{equation}

Let us say $f(\theta_{a})$ has parameters $\kappa$ and $\mu$, and $f(\theta_{ab})$ has parameters $\bar{\kappa}$ and $\bar{\mu}$. Then, according Eqn. (\ref{equation:dkl}), it has
\begin{equation}
\begin{split}
      D_{\mathrm {KL} }[f(\theta _{a})\|f(\theta_{ab})] = & {\frac {1}{2}}\sum _{jk}\Delta \theta ^{j}\Delta \theta ^{k} f_{jk}(\theta _{a}) \\
      = &{\frac {1}{2}}[\tau_{\kappa} (\bar{\kappa},\bar{\mu})(\bar{\kappa} - \kappa)^{2} + 2\tau_{\kappa\mu} (\bar{\kappa}) \bar{\mu}^{T} 
      (\bar{\kappa} - \kappa) (\bar{\mu}-\mu) \\
      +& \tau_{\mu} (\bar{\kappa}) (\bar{\mu} - \mu) 
      \bar{\mu} \bar{\mu}^{T} (\bar{\mu} - \mu) 
      ].
\end{split}
\end{equation}
\par After constructing CSG, we can get an adjacency matrix $A$. Non-zero values in the adjacency matrix indicate the existence of links between nodes. The values are normalized, and the sum of each row or column is equal to 1.

\subsection{Graph convolution network}
CSG is highly valuable to identify the nodes. To leverage this, we adopt a graph convolution network (GCN) to perform reasoning on CSG. Specifically, in order to adapt to the changing person features distribution caused by the open-world challenge, we use GCN to project the features into a linearly separable low dimensional space. With a simple Nearest Neighbor strategy adopted after then for the discrimination, the method can achieve high precision clustering results.

The input of the GCN is the original node feature matrix together with the adjacency matrix $A$. The output is the projected node feature matrix with the same number of input features but with lower dimensions. Each graph convolution layer is inputted by the last node feature matrix together with the adjacency matrix and outputs the next node feature matrix. Formally, for the $l^{th}$ layer, we have the following equation:
\begin{equation}
     \mathbf{X}^{l + 1} = \sigma([\mathbf{X}^{l} | \mathbf{\Lambda}^{-\frac{1}{2}} \mathbf{A} \Lambda^{-\frac{1}{2}} \mathbf{X}^{l} ] \mathbf{W}_{l}),   
\end{equation}
where $\mathbf{X}^{l} \in \mathbb{R}^{n \times d_{l}}$ is the node feature matrix inputted to the $l^{th}$ layer. $n$ represents the number of nodes, and $d^l$ represents the feature's dimension of the $l^{th}$ layer. $\mathbf{\Lambda}$ is a diagonal matrix with $\mathbf{\Lambda}_{ii} = \sum\nolimits_{j} \mathbf{A}_{ij}$. $\sigma(\cdot)$ is an nonlinear activation function. $\mathbf{W}_{l} \in \mathbb{R}^{2d_{l} \times d_{l+1}}$ is a layer-specific trainable weight matrix for the convolution layer. $\mathbf{X}^{l+1} \in \mathbb{R}^{n \times d_{l+1}}$ is the output node feature matrix of the layer.
\par GCN is supervised to project the nodes of the same person to be closer and others to be more distant. Toward that end, We adopt triplet loss \cite{schroff2015facenet} to our case. Denote the outputted features matrix as $\mathbf{O} \in \mathbb{R}^{n \times d_{out}}$, and each feature for one node as $o \in \mathbb{R}^{d_{out}}$. Given a crowded graph containing $h$ people, and each person corresponding $c$ node in the graph, we denote the feature of $k^{th}$ node of $i^{th}$ person as $o_{i}^{k}$, then our loss is given by:
\begin{equation}
\begin{split}
    \mathcal{L}_{GCN} = \sum_{i = 1}^{h} \sum_{k=1}^{c_{i}} \max(log \sum_{r=1,r \neq k}^{c_{i}} e^{D(o_{i}^{k} , o_{i}^{r})} 
    + log \sum_{j=1, j \neq i}^{h} \sum_{s = 1}^{c_{j}} e^{m- D(o_{i}^{k} , o_{j}^{s})},0),
\end{split}
\end{equation}
where $m$ is the least margin that the projected feature is closer to the same class than the different classes. $D$ is the distance measure, which is implemented by cosine similarity here. 

In the inference stage, we use a simple Nearest Neighbor (NN) clustering strategy on the projected features. The NN has two types, the NN-A (Assign a new \textit{query} to the existing nodes) and NN-M (Merge existing nodes). We set different distance thresholds $\xi_{A}$ and $\xi_{M}$ for the NN-A and NN-M, respectively. And the distance measure is set to be as same as that in the training stage. NN-A will assign the new \textit{query} to the nearest \textit{person records} if their distance is lower than $\xi_{A}$. If this distance is alternatively higher than the threshold $\xi_{A}$, the \textit{query} will be left as an outlier, which represents a new \textit{person record}. The outliers will be processed as nodes when being contained in the other CSG.

NN-M merges two \textit{person records} together if their distance is lower than $\xi_{M}$, which is used to make up the early stage division problem (predict one customer as multiple people). In this data-stream clustering process, with the continuous arriving of the queries, a node has the chance to be actives in many different sub-graphs. This process helps to gradually attach the global information, then correct the previous errors caused by over-conservative clustering strategy.
\section{Experiments}
\subsection{Data and Evaluation metric}

\newcommand{\tabincell}[2]{\begin{tabular}{@{}#1@{}}#2\end{tabular}}  

\begin{table}[!t]
	\centering
	\caption{Ablation study setup. 'vMF' and 'Cos' represent applying vMF similarity and Cosine similarity in the corresponding element, respectively. The 'NN-A' represents Nearest Neighbor clustering which Assigns a \textit{query} to existing nodes. The 'NN-M' represents Nearest Neighbor clustering which Merges existing nodes}

	\begin{tabular}{c|ccccc}
		\hline
        \diagbox[]{Model}{Element}   & \tabincell{c}{Time-Space\&Track} & CSG & GCN & NN-A &  \tabincell{c}{NN-M}  \\ \hline

        Baseline    & -                  &-            &-   & Cos  &- \\
        TS    &$\surd$           &-             &-    & Cos  &-\\
         TS-M    &$\surd$            &-            &-     & Cos  & Cos\\
     TS-M-vMF    & $\surd$           &-           &-     & vMF  & vMF\\
     \hline
     Cos-GCN   & $\surd$            & Cos       & $\surd$  & Cos  & Cos\\
        CSG-GCN   & $\surd$            & vMF      & $\surd$  & Cos & Cos \\
		\hline
	\end{tabular}

	\label{tabel:ablation_setup}
\end{table}

\begin{table}[!t]
	\centering
	\caption{Ablation study results. $P$ : BCubed Precision, $R$ : BCubed Recall, $F=\frac{2PR}{P+R}$, $T$ : Time (Seconds per one thousand quires)}
	\begin{tabular}{c|cccc|cccc}
		\hline
		& \multicolumn{4}{c|}{DaiCOFG} & \multicolumn{4}{c}{IseCOFG} 
		\\ \hline
		& $P$ & $R$ & $F$ & $T$ & $P$ & $R$ & $F$ & $T$\\ \hline
		Baseline  & 84.46 & 73.28 & 78.47 & 46.61  & 86.62 & 84.21 & 85.40 & 26.78 \\ 
		TS  & 91.25 & 80.16 & 85.35 & \textbf{8.27}  & 93.31 & 84.02 &  88.42 & \textbf{7.32} \\ 
		TS-M  & 91.22 & 84.39 & 87.67 & 9.13 & 96.02 & 89.30 & 92.54 & 7.95  \\ 
		TS-M-vMF  & 95.75 & 93.70 & 94.71 & 11.06  & 97.47 & 94.51 & 95.97 & 8.72 \\ 
		Cos-GCN & 96.83 & 96.37 & 96.60  & 11.81 & 97.73 & 96.91 & 97.32 & 8.96 \\ 
		CSG-GCN  & \textbf{98.77} & \textbf{98.24} & \textbf{98.50} & 13.68 & \textbf{99.07} & \textbf{98.75} & \textbf{98.91} & 10.54 \\ 
		\hline
	\end{tabular}
	\label{tabel:ablation}
\end{table}

\begin{table}[!t]
	\centering
	\caption{Comparison with SOTA. $P$ : BCubed Precision, $R$ : BCubed Recall, $F=\frac{2PR}{P+R}$, $T$ : Time (Seconds per one thousand quires, omitted if $>$ 50 s/ptq)}
	\begin{tabular}{c|cccc|cccc}
		\hline
		& \multicolumn{4}{c|}{DaiCOFG} & \multicolumn{4}{c}{IseCOFG} 
		\\ \hline
		& $P$ & $R$ & $F$ & $T$ & $P$ & $R$ & $F$ & $T$\\ \hline
		Scalable-kmeans  & 84.49 & 79.36 & 81.84 & 16.68  & 86.63 & 84.36 & 85.48 & 16.15 \\ 
		ClusTree  & 91.81 & 85.73 & 88.67  & 14.00 & 93.41 & 90.64 & 91.87
 & 14.00 \\ %
		\hline
		ClusterGCN & 94.55 & 90.75 & 92.61  & - & 94.12 & 91.46 &  92.77 & - \\ 
		AffinityGCN & 97.05 & 96.82 & 96.93  & - & 98.47 & 97.83 &  98.14 & - \\ 
		GraphSaint & 95.84 & 92.34 & 94.05  & 46.62 & 96.16 & 95.67 &  95.91 & 28.79 \\ 
		IPS-GCN  & 96.32 & 94.17 & 95.23  & \textbf{11.49} & 97.15 & 96.30 &  96.72 & \textbf{9.69} \\ 
		GLCN  & 96.86 & 94.96 & 95.90  & 12.37 & 97.82 & 96.51 & 97.16 & 10.37 \\ \hline
		CSG-GCN  & \textbf{98.77} & \textbf{98.24} & \textbf{98.50} & 13.68 & \textbf{99.07} & \textbf{98.75} & \textbf{98.91} & 10.54 \\ 
		\hline
	\end{tabular}
	\label{tabel:comparision}
\end{table}

It should be noted that all the customers we collected information from assigned the informed agreement before going into the grocery. In addition, all data points used in the clustering stage are obfuscated high-dimensional features, which do not contain any customers' personal information.

To evaluate our method in different scenes, we establish two datasets collected from a large grocery and a smaller grocery, respectively. The recurring identities across training and test set are removed to avoid bias. 
The dataset we collected from a large grocery is called DaiCOFG. DaiCOFG contains $362,300$ snapshots with $10,176$ identities for training, in which $125,378$ snapshots are labeled, and each identity contains at least one snapshot labeled. DaiCOFG contains $250,710$ labeled snapshots with $7,406$ identities for testing. The snapshots are taken by 186 cameras deployed at the key spots of the grocery. The dataset we collected from the smaller grocery is called IseCOFG. IseCOFG contains $78,630$ snapshots with $4,116$ identities for training, in that $21,648$ snapshots are labeled, each identity contains at least one snapshot labeled. IseCOFG contains $54,606$ snapshots with $2,773$ people for testing. The snapshots are taken by 76 cameras in the grocery. 

To evaluate the performance of the proposed algorithm, we adopt the mainstream BCubed evaluation metrics \cite{Enrique2009A,wang2019linkage}. Denote ground truth label and predicted label as a $y$ and a $y'$ respectively, the pairwise correctness is represented as:
\begin{equation}
 Correct(i,j)=\left\{
\begin{array}{rcl}
1       &      & {y_i = y_j \ and \ y'_i = y'_j}\\
0       &      & {otherwise}\\
\end{array} \right.,
\end{equation}
\par If the $i^{th}$ \textit{query} and the $j^{th}$ \textit{query} belong to the same customer during clustering and labeling, we can get $Correct(i,j)=1$. The BCubed Precision $P$ and BCubed Recall $R$ are respectively defined as:
\begin{equation}
\begin{split}
    P = \mathbb{E}_{i} [\mathbb{E}_{j:y'_{i} = y'_{j}}[Correct(i,j)]], \ 
    R = \mathbb{E}_{i} [\mathbb{E}_{j:y_{i} = y_{j}}[Correct(i,j)]],
\end{split}
\end{equation}
\par When taking both precision and recall into consideration, BCubed F-measure is defined as $F = \frac{2PR}{P+R}$.To evaluate the algorithms' speed, we record the seconds per one thousand quires (s/ptq) as the metric in the comparisons. 

\subsection{Experiment Setting}

The variables in \textbf{link} operation are pre-trained on the training set of the selected dataset of the experiment. In the training stage, it is supervised by binary cross-entropy loss function with a mini-batch of 64 for 80 epochs using ADAM algorithm \cite{kingma2014adam}. The learning rate is set to 0.01. We set the number of convolution layers in our GCN to 3. The number of units in the graph convolution network's hidden layer is set to 256, 128, and 64. The number of the nearby nodes is set as 256. We train GCN for a maximum of 120 epochs (training iterations) using an ADAM algorithm with a learning rate 0.01, and stop training if the validation loss does not decrease for 10 consecutive epochs, as suggested in work \cite{kipf2016semi}. All the network weights $\theta$ are initialized using Glorot initialization \cite{glorot2010understanding}. The thresholds $\xi_{A}$ and $\xi_{M}$ are set as 0.91 and 0.88, respectively. 

The experiments are run on the server cluster with 16 CPU: Intel Xeon Gold 5120, 256GB memory, and 8 GPU: NVIDIA Tesla P40. 

\subsection{Ablation Study}
To show the advantages of the proposed components, we do comprehensive ablation studies. The setup and results of the ablation study are shown in Table \ref{tabel:ablation_setup} and Table \ref{tabel:ablation}.

\par As shown in Table \ref{tabel:ablation}, the raw NN-A strategy based only on the features' cosine similarity (Baseline) takes a long time and gets a lower recall. Applying time-space constraint and track information (TS) helps to narrow the search range since we only compare the nodes under time-space constraints and linked through track information, which speeds up the algorithm a lot with 6.88\% and 3.02\% F1 score improvement on DaiCOFG and IseCOFG respectively.
Applying NN-M strategy (TS-M) actually turns the method to a k-means liked algorithm. It can be seen that, NN-M strategy helps to significantly improve the recall with a slight degradation of the precision. 
Further applying vMF-based divergence in NN-A and NN-M (TS-M-vMF) significantly improves both the precision and the recall. Also, the processing time increased by the extra consumption. Applying GCN (Cos-GCN) helps to improve the recall, due to the low-dimensional features can be better discriminated in the early stage. Further replacing cosine similarity by vMF-based divergence to construct the graph (Cos-GCN$\rightarrow$CSG-GCN) turns it to the proposed algorithm. It can be seen that vMF weight strategy improves 1.90\% and 1.59\% F1 score on DaiCOFG and IseCOFG, respectively.

From the comparison of TS-M/Cos-GCN with TS-M-vMF/CSG-GCN, it can be seen that modeling the person features by vMF distribution significantly outperforms cosine similarity based pointwise comparison. Comparing TS-M with Cos-GCN, the combination of proposed CSG and GCN shows better precision and recall even without vMF distribution modeling. This shows the effectiveness of the proposed CSG when facing the dynamic and unseen person flow. In general, from Table \ref{tabel:ablation}, we can see that the proposed algorithm CSG-GCN gets both high performance and processes in the tolerable time on the dataset.
\subsection{Comparison with alternative methods}
In this part, to verify the effectiveness of the proposed method, we compare the proposed method with a wide range of alternative clustering methods, including non-learning-based and learning-based methods. Since those methods are not designed for our scene, and most of the learning-based methods are oriented to closed data sets, we have to adapt some methods for the comparison.

\par Most previous methods toward data-stream clustering are not learning-based. We compare our method with scalable k-means \cite{bradley1998scaling} and ClusTree \cite{kranen2011clustree}, which are two commonly used methods in the data-stream clustering tasks. Scalable k-means employs different mechanisms to identify objects that need to be retained in memory. It stores data objects in a buffer in the main memory. By utilizing CF vectors, it discards objects that were previously statistically summarized into the buffer. When the block is full, an extended version of k-means is executed over the stored data. ClusTree algorithm \cite{kranen2011clustree} proposed to use a weighted CF vector, which is kept into a hierarchical tree (R-tree family). ClusTree provides strategies for dealing with time constraints for anytime clustering, that is, the possibility of interrupting the process of inserting new objects in the tree at any moment.  

Recently, GCN has been proved an efficient method for clustering tasks \cite{yang2019learning}. However, few works applied GCN to the data-stream clustering. We adapted ClusterGCN\cite{clustergcn}, AffinityGCN\cite{affinitygcn}, GraphSaint\cite{graphsaint},  IPS (Instance Pivot Subgraph)-GCN\cite{wang2019linkage} and GLCN\cite{Jiang_2019_CVPR} for the comparison. The results are shown in Table. \ref{tabel:comparision}. \cite{wang2019linkage,clustergcn,affinitygcn,graphsaint} cluster the features by applying GCN on the constructed subgraph. Their methods are applied to construct the subgraph for each coming \textit{query} and then trained by our own strategy. \cite{Jiang_2019_CVPR} proposed to learn a graph for GCN clustering based on Euclidean distance, which is called Graph Learning Convolutional Network (GLCN). To adapt it to our data-stream scenario, we apply it on CSG linked nodes to cluster each sub-graph locally. 

Since the algorithm takes more time when the grocery becomes larger, we record the algorithms' speed when the grocery is full. ClusTree is a time-adaptable method that fits our scenario well. However, when we simulate the fast stream in the open checkout-free groceries (e.g., 14s/ptq), ClusTree gets lower Precision and recall.

In addition, through Table \ref{tabel:comparision}, we can see that learning-based methods are much more efficient than traditional methods. They generally achieve better performance by projecting the features to the low-dimensional linear subspace. \cite{affinitygcn} achieves competitive overall performance due to the global-aware clustering, but the time consumption of the strategy is intolerable in this scene. IPS based GCN\cite{wang2019linkage,Jiang_2019_CVPR} are more efficient comparing with the others. To take our \textit{query} as pivot for IPS construction, the sub-graph represents the local correlation of nodes just like the proposed method, and gains a balanced time and performance improvement. However, they still measure the node distance by cosine similarity, which ignores the distribution nature of the nodes.

As shown in Table \ref{tabel:comparision}, the proposed method surpasses the other clustering methods by a large margin and achieves state-of-the-art performance on the datasets. It also outperforms CSG-linked GLCN by a 2.60\% and 1.75\% F1 score on DaiCOFG and IseCOFG, respectively, with comparable time consumption, indicating the vMF-based \textit{weight} strategy works even better than the learning-based strategy. In practice, the proposed method is thus the most applicable algorithm for the complicated open check-out free grocery scenario. 
\section{Conclusion and future work}
This paper proposed a real-time Person Clustering method, namely CSG-GCN, for Open checkout-free groceries under data streams and the unknown number of persons. The proposed method fully utilizes the human time-space information and makes a variance-considered comparison on the spherical summarized data to improve the method's speed and accuracy. And the experimental results show the effectiveness and advantages of the method. Although this strategy is efficient for the inference, it is in fact time and memory consuming in the training. Future research can focus on designing the more training-friendly strategies for this task. Moreover, the proposed method consists of several, relative individual components, which causes the inconvenience in the application, e.g., the difficulties in hyper-parameters picking. We are looking forward a simpler and more consistent solution for this task.

\clearpage
%
%
\bibliographystyle{splncs04}
\bibliography{egbib}
\end{document}